\documentclass[12pt]{ieeeconf}

\IEEEoverridecommandlockouts
\pdfoutput=1
\usepackage{comment}
\usepackage[utf8]{inputenc} 
\usepackage[T1]{fontenc}    
\usepackage{hyperref}       
\usepackage{etoolbox}
\usepackage{cuted}
\usepackage{flushend}
\AtBeginEnvironment{tabular}{\small}
\hypersetup{
	colorlinks=true,
	linkcolor=blue,
	citecolor=blue,
	urlcolor=blue,
}
\usepackage{url}            
\usepackage{booktabs}       
\usepackage{amsfonts}       
\usepackage{nicefrac}       
\usepackage{microtype}      
\usepackage{lipsum}
\usepackage{footmisc}
\usepackage{multicol}

\usepackage{slashbox}
\usepackage[libertine]{newtxmath}
\usepackage{color}
\usepackage{setspace}
\usepackage{xcolor}
\usepackage{pdflscape}
\usepackage{graphicx} 
\usepackage{float}
\usepackage{booktabs}
\usepackage{algorithm}
\usepackage{algorithmicx}
\usepackage{algpseudocode}	
\usepackage{indentfirst}
\usepackage{xltabular}
\usepackage{array, makecell} %
\usepackage{xcolor}
\usepackage{pgfplots}
\usepackage{tikz}
\usepackage{subcaption}
\usepackage[utf8]{inputenc}
\usepackage{authblk}
\usepackage{graphicx}
\usepackage{fixltx2e}
\usepackage{booktabs}
\usepackage{array}
\graphicspath{ {./Pix/} }
\renewcommand{\arraystretch}{1.4}

\definecolor{bblue}{HTML}{4F81BD}
\definecolor{rred}{HTML}{C0504D}
\definecolor{ggreen}{HTML}{9BBB59}
\definecolor{ppurple}{HTML}{9F4C7C}
\definecolor{yyellow}{HTML}{FFD700}
\definecolor{ppink}{HTML}{FE6F5E}
\definecolor{purpule}{HTML}{BF94E4}

\title{\LARGE \bf Meta-SeL: 3D-model ShapeNet Core Classification \\ using Meta-Semantic Learning}

\author[1]{Farid Ghareh Mohammadi*}
\author[1]{Cheng Chen}
\author[1]{Farzan Shenavarmasouleh}
\author[2]{\\M. Hadi Amini}
\author[1]{Beshoy Morkos}
\author[1]{Hamid R. Arabnia}
\affil[1]{Department of Computer Science, University of Georgia, Athens, GA, USA \authorcr {\{farid.ghm, fs04199,cheng.c,bmorkos, hra\}@uga.edu}\vspace{1.5ex}}
\affil[2]{ Knight Foundation School of Computing and Information Sciences,\protect\\ Florida International University, Miami, FL, USA\authorcr {\{moamini\}@fiu.edu}\vspace{1.5ex}}

\begin{document}
\maketitle 
\thispagestyle{empty}
\pagestyle{empty}
\begin{abstract}
Understanding 3D point cloud models for learning purposes has become an imperative challenge for real-world identification such as autonomous driving systems. A wide variety of solutions using deep learning have been proposed for point cloud segmentation, object detection, and classification. These methods, however, often require a considerable number of model parameters and are computationally expensive. We study a semantic dimension of given 3D data points and propose an efficient method called Meta-Semantic Learning (Meta-SeL). Meta-SeL is an integrated framework that leverages two input 3D local points (input 3D models and part-segmentation labels), providing a time and cost-efficient, and precise projection model for a number of 3D recognition tasks. The results indicate that Meta-SeL yields competitive performance in comparison with other complex state-of-the-art work.
Moreover, being random shuffle invariant, Meta-SeL is resilient to translation as well as jittering noise.
\end{abstract}

\section{Introduction}

\parindent0pt$\diamond$\textbf{Motivation}
Analyzing 3D point cloud models has become an essential component of many AI systems such as additive manufacturing \cite{lyu2021situ}, autonomous driving \cite{kidono2011pedestrian}, and augmented reality \cite{alexiou2017towards}. Classifying 3D point clouds accurately in real-time can be a challenging task due to the complexity of objects that exist in a real-world environment \cite{uy2019revisiting}. Among various 3D models in computer vision, point cloud representations have become increasingly popular as more data is collected by devices such as LiDAR. The goals of modeling 3D objects can be divided into three categories: part segmentation, semantic segmentation, and shape classification. Literature shows promising results utilizing sophisticated PointNet-based deep learning techniques for understanding object classifications from point clouds \cite{uy2019revisiting,liu2019relation}. These methods, however, often require a considerable number of model parameters and are computationally expensive. In this study, we present a framework for simplifying the classification process by executing one epoch while achieving competitive results. 

\begin{figure*}
	\centering
		\includegraphics[height =8em]{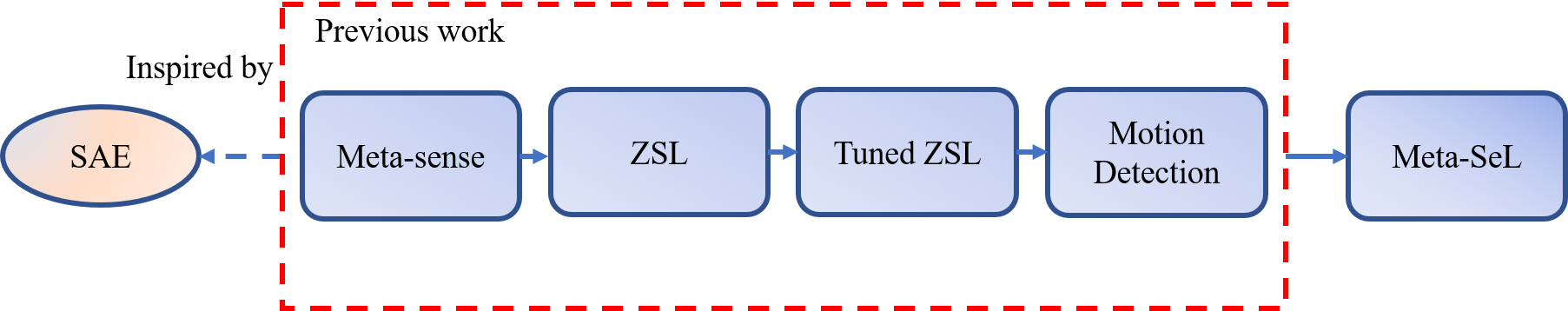}
	\centering
	\caption{The progress of the work from Meta-sense to Meta-Semantic Learning (Meta-SeL)}
	\label{fig:history}
\end{figure*}

\parindent0pt$\diamond$\textbf{Gap:} One major challenge in classifying 3D models is to improve algorithm efficiency and execution in real-time. Several deep neural networks were developed to address this issue, including PointNet, DGCNN, and SimpleView. With a unified architecture, PointNet utilizes permutation invariance of points and processes each independently with a symmetric function that aggregates features. DGCNN introduces the EdgeConv block, which exploits both local and global shape properties for each point as topological information. In SimpleView, 3D point clouds are converted into 2D depth images using a projection-based method as a type of dimension reduction technique. In general, deep neural models require a long training time and rely on many parameters to reach higher accuracy. As yet, these techniques do not convert feature space into semantic embedding space for the problem of 3D model classification.

\parindent0pt$\diamond$\textbf{State of the art solutions:} Recently 3D point cloud datasets and associated solutions for segmentations and classifications are presented and the need to conduct research on 3D Models is proliferated \cite{qi2017pointnet,xie2021generative,huang2021metasets}. Here, we summarise the most popular ones.  The most common benchmarks  for comparing methods
for point cloud classification are ShapeNet Core \cite{yi2016scalable} and Modelnet40 \cite{wu20153d}. There are current solutions tested on  ModelNet40,  such as Point-Net \cite{qi2017pointnet}, DGCNN \cite{DGCNN2019} and RSCNN \cite{liu2019relation}. In this research, we apply two of these current solutions, DGCNN and Point-Net, with which we compare the results.

\vspace{2pt}$\noindent\diamond$\textbf{Contribution:}
In this paper, we propose an efficient solution to classify 3D point cloud models that is highly competitive in comparison with other state-of-the-art methods. It is also efficient in terms of resource utilization. Furthermore, we develop a simple framework, Meta-SeL, to perform linear computations on CPUs rather than GPUs. We originally, as shown in Figure \ref{fig:history}, proposed the idea of Meta-sense in \cite{mohammadi2019promises}, to perform a zero-shot learning (ZSL) task \cite{mohammadi2020introduction}, and, further, we presented another work on computer vision using tuned-ZSL \cite{mohammadi2019parameter}, and another on signal processing for human motion detection using ZSL \cite{mohammadi2021human}.

We conduct this research based on previous work mentioned, inspired by one part of the semantic auto-encoder solution presented initially by Kodirov et al. \cite{kodirov2017semantic} in which we generate a projection function (\textit{W}) from vector space to a semantic space. The key contributions of our research in this collaborative work are as follows:

\parindent10.5pt$\diamond$ Design a novel framework, namely Meta-SeL, suitable for 3D point cloud models.

$\diamond$ Posit that training and testing phases only need one \textit{epoch} to yield a promising result. This means that Meta-SeL has a high convergence rate, so it is extremely time and cost-efficient.  

$\diamond$ Demonstrate how Meta-Sel has a large positive impact on the point cloud, particularly shapeNet Core dataset, classification performance.  

$\diamond$ Demonstrate that Meta-SeL is resistant to jittering noise and translation.


$\diamond$ Illustrate that Meta-SeL generates semantic information for each entry on the fly.

The reminder of this paper is organized as follows. In section 2, we present an overview of the current solutions for 3D model classification and their learning process. The proposed method, Meta-SeL, is detailed in section 3. Section 4 reports on experimental results, discussion and analysis.

 \begin{figure*}
	\centering
		\includegraphics[width =0.85\textwidth]{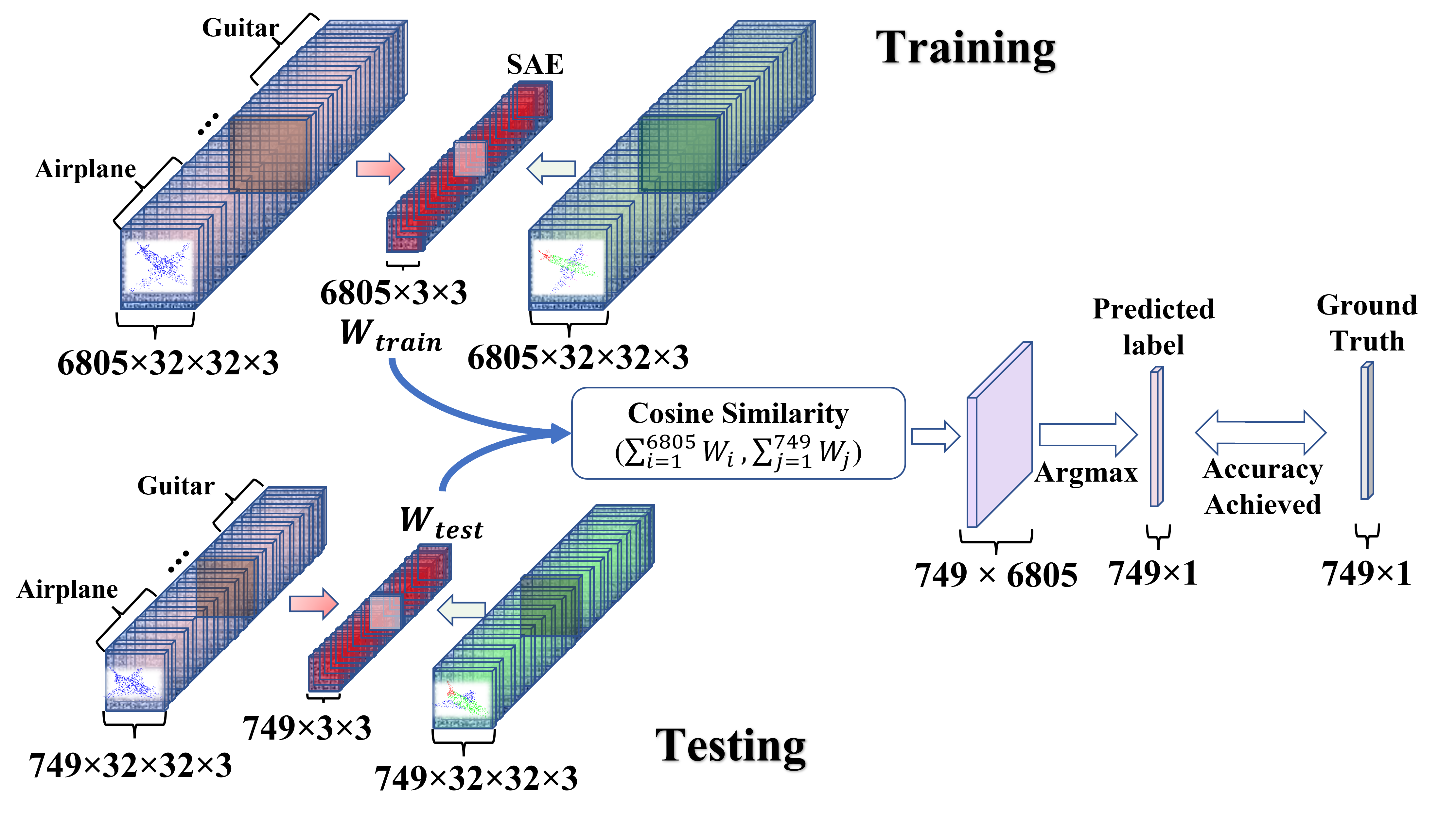}
	\centering
	\caption{Meta-SeL framework for 3D point cloud models classification}
	\label{fig:Meta_SeL}
\end{figure*}

\section{Related Work}

Researchers investigated  3D point clouds models to understand and analyze them using handy-crafted features \cite{aubry2011wave,rusu2008aligning}. These features are extracted and categorized into two parts: local features and global features. Researchers feed these features into traditional machine learning algorithms to obtain results. However, these features are biased and may not work for all 3D model shapes. To address this challenge, scientists use deep learning with PointNet \cite{qi2017pointnet} in 2017 and further PointNet$++$ \cite{qi2017pointnetplusplus}. So, deep learning has been widely used in 3D model classification with the base of PointNet such as DensePoint \cite{liu2019densepoint}. We discuss the 3D dataset and the difference between deep learning algorithms such as PointNet and DGCNN.
  
    $\noindent \diamond$\textbf{3D Point Cloud Data:}
    A subset of the ShapeNetCore dataset is created to propagate manually verified labels from a large dataset to a small dataset \cite{yi2016scalable}, which includes 17,775 models from 16 categories. Each model is annotated in two to six parts. The dataset is initially divided into 15990 and 1785 models, respectively, for training and testing. 

    $\noindent\diamond$\textbf{Approaches:} PointNet has revolutionized the process of classifying and segmenting 3D pointcloud datasets \cite{qi2017pointnet}. There are three appealing properties of this approach: unordered input, local and global feature extraction, and invariance under transformations. Through a local network, i.e. T-net, spatial information has been transformed into local and global features. In contrast to PointNet, DGCNN \cite{DGCNN2019} incorporates structural information by computing the pairwise distance using k-nearest neighbors(KNN) of points.
    
    $\noindent \diamond$\textbf{Meta Learning:}
     Meta-learning has three main promises to classify objects and images. One of which is zero shot-learning \cite{mohammadi2020introduction}, in which we learn from seen classes to predict unseen classes. Meta-sense \cite{mohammadi2019promises}, image recognition \cite{mohammadi2019parameter}, and human motion recognition \cite{mohammadi2021human} are a few examples to present the process of generating a projection function from a vector space to a semantic space. These research studies are inspired by semantic auto-encoder (SAE) \cite{kodirov2017semantic}.

\section{Proposed Method}
 
 Since the number of 3D point cloud models classification algorithms has increased, 
 an efficient method to perform this task must be developed. To that end, we present an optimized framework leveraging part-segmentation label information to classify 3D points cloud models. Figure \ref{fig:Meta_SeL} presents a comprehensive framework of Meta Semantic Learning. We start with point cloud and elaborate on Meta-SeL. Meta-SeL is an extended and optimized version of zero shot learning that works for 3D model classification. The code is available at github \footnote{\url{ https://github.com/faridghm/Meta-SeL.git}}
 
 \vspace{1pt} $\noindent \diamond$\textbf{Point Cloud:}    To leverage prior information on both classification and segmentation labels, we select the models with three segmentation labels. As shown in Table \ref{table2}, the total number of models has reduced to 7,555 with 10 categories.  Note that each filtered model contains three parts of segmentation labels: 1, 2, and 3 which cover all the point cloud model.

 \begin{figure*}
 
		\includegraphics[height=28em]{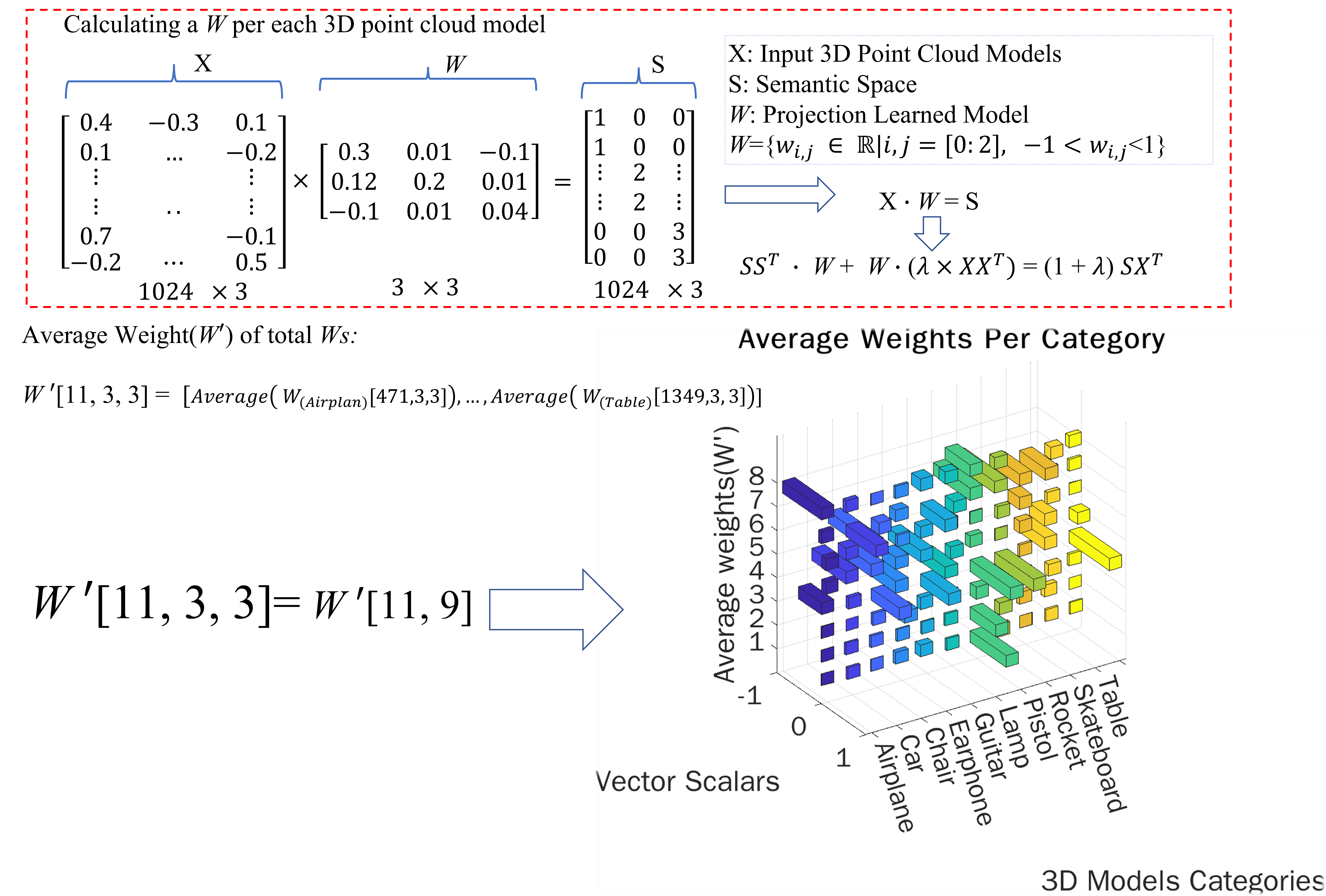}
	\centering
	\caption{Averages Weights (\textit{W}s)}
	
	  \label{fig:sub2}
 
\caption{Presents the process to generate \textit{W}s, together with average weight \textit{W'} per each category. } 
\end{figure*}

 \vspace{1pt}$\noindent\diamond$\textbf{Pre-Processing:}  Before starting Meta-SeL process, we have to make sure all input 3D could point models include three-part segmentation labels. Each model has a dimension of [32 * 32 * 3] and associated part-segmentation labels 
 which stand for 1,2, or 3. We take advantage of the three-dimensionality of models and present a framework to achieve promising results for model classification. Thus, we maintain models that have a number of part segmentation labels equal to 
 three. Next, we convert the input models (models and segmentation labels) into a new dimension 1024*3 models. Finally, we sort segmentation labels in ascending order and update the main models accordingly to have a consistent order of part-segmentation labels for all input 3D point cloud models. Note that we assume that every point cloud in the 3D point clouds models is independent.
  
\vspace{1pt}$\noindent\diamond$\textbf{Latent space (semantic space):}
Latent space represent a compressed version of input data in a way that helps decoders to generate an output and predict a label to a given sample instance. In prior research studies, scientists used a static semantic space similar to \cite{kodirov2017semantic} or leveraged auto-encoders to generate the semantic space using deep neural network. But they made a static general semantic space for all input data.

Unlike other semantically based algorithms, Meta-SeL performs a classification task with dynamic semantic space for each 3D model separately. Using consistent ordering of the associated part-segmentation labels, Meta-SeL generates a semantic space for each new 3D model instantaneously. 
Each column in the semantic space matrix(S) represents the semantic space for cloud points. We use this dimension [32*32*3]  consistent with the input 3D model dimensions and size. 

\vspace{1pt}$\noindent\diamond$\textbf{Projection Learned Model:}
Projection Learned Model(PLM) plays a significant role in Meta-SeL during the process of understanding input 3D point cloud  models. 

We generate PLM \textit{W}s in both phases: training \textit{$W_{train}$} and testing \textit{$W_{test}$} separately in real-time while executing each step. We are inspired by the idea of the semantic auto-encoder \cite{kodirov2017semantic,mohammadi2021human,mohammadi2019parameter} in which researchers applied the framework of an auto-encoder to generate a projection function (\textit{W}) to map input vectors to a latent space. To generate the \textit{W}, we use only the encoder part of the semantic auto-encoder to generate a project function for each model. The encoder uses the Sylvester equation to find \textit{W} 
for each input 3D model. Figure \ref{fig:sub2} presents a conceptual view of the projection learned model generation where 
the Sylvester equation takes input data(X,S and$\lambda$)  to generate projection learned models (\textit{W}s) for both train and test.
Thus, the output of the function is \textit{W}s with associated labels. Then, we calculate an average weight (\textit{W'}) of each category to present distinct distributions of categories.
It is obvious that Meta-SeL generates complete separate \textit{W}s that enables us to predict test 3D models accurately and precisely. We have three dimensions that are shown in figure \ref{fig:sub2}. First, On the x-axis we have the 3D model categories. Second, we have associated average weights (\textit{W'}s), which are flattened into 1*9 on z-axis. Finally, on Y-axis we have scalar values for the
\textit{W'}s. 

After the training phase is successfully completed, we execute the testing phase and evaluate the Meta-SeL results using a cosine similarity metric. We use this metric to seek the closest model to the test model. 
The cosine similarity metric seeks the angles between two vectors rather than their weights. Thus, if two models are far apart from each other, this metric can still  determine if they are similar or not using equation of cosine similarity. The formula for the cosine similarity is given in equation \ref{cos-Sim}.

\begin{figure*}
	\centering
		\includegraphics[height=0.65\textwidth]{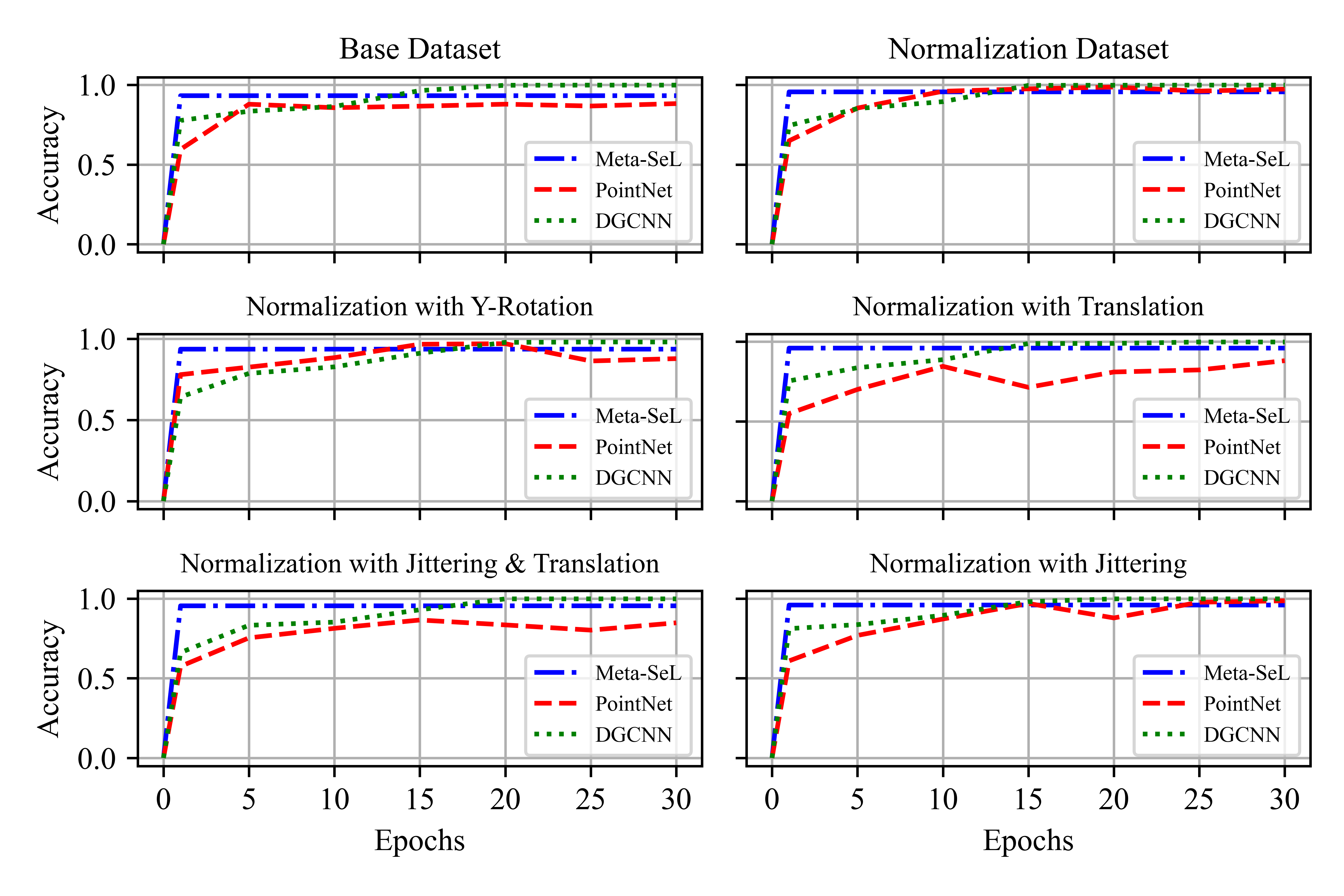}
	\caption{Accuracy Comparison}
	  \label{fig:sub1}
\end{figure*}

$\noindent\diamond$\textbf{Accuracy Computation:} We use an arg-max function to predict the test 3D model labels using a calculated cosine-similarity matrix and compare them with the ground truth to compute the accuracy of Meta-SeL. Meta-SeL computes the final result with only one \textit{epoch} and does not require further iterations, as its accuracy does not change. Figure \ref{fig:sub1} presents Meta-SeL evaluation performance with DGCNN and PointNet on different types of customized and augmented datasets. We observe that the accuracy of Meta-SeL in the beginning reaches the highest rate and remains there.

  \begin{strip}
  \begin{equation}
    \label{cos-Sim}
\end{equation}
  \LARGE
\centering

$\cos ({\bf W_{train}},{\bf W_{test}})= {{\bf W_{train}} {\bf W_{test}} \over \|{\bf W_{train}}\| \|{\bf W_{test}}\|} = \frac{ \sum_{i=1}^{n}{{\bf W_{train}}_i} \sum_{i=1}^{n}{{\bf {\bf W_{test}}_i} }}{ \sqrt{\sum_{i=1}^{n}{({\bf W_{train}}_i)^2}} \sqrt{\sum_{i=1}^{n}{({\bf W_{test}}_i)^2}} }$
 
\end{strip}

    \newcommand{\tol}[3]{\ensuremath{\si{#1}^{+\num{#2}}_{-\num{#3}}}}
  
    \begin{table*}[!ht]
    \caption{a. Presents different results of Meta-SeL on 3D model Classification in comparison with other state of the art methods b. Presents all class distribution during training and testing phase, together with \textit{recall} and \textit{precision} c. Meta-SeL's Confusion matrix for normalized dataset }
      \begin{subtable}{.4\linewidth}
    \centering
    \caption{ Results comparison with other state of the art }
    \label{table1}
      \begin{tabular}{l|c}\toprule
     Methods & Accuracy($\%$)   \\ \midrule
    DGCNN: & 99.83 \\ 
    PointNet: & 88.22  \\ 
    \bottomrule
    \textbf{Meta-SeL:}  &  \\
     Base :   & 93.19 \\
     Random Shuffle(RS) & 93.19 \\
    Sorted(Ascending) & 93.19 \\
     Sorted(Descending) & 93.19 \\
     $Normalization (N) $ & \textbf{95.59}\\
    $N + RR (x-axis)$ & 90.25\\
     $N + RR (y-axis) $ & 93.59 \\
     $N + RR (z-axis) $ & 83.97\\ 
     $N + RR (x, y, z )$ & 88.38\\ 
     $N + T $ &  \textbf{95.99}\\
     $N + RR + T $ & 77.03\\ 
    $N + J $ &  \textbf{95.95} \\
     $N + J + T $&  \textbf{95.46}\\
    
    \bottomrule
      \end{tabular}
 \end{subtable}%
    \begin{subtable}{.4\linewidth}
    \centering
    \caption{Class distribution in the dataset}
    \renewcommand{\arraystretch} {1}
    \hspace{-1cm}\begin{tabular}{c|c|c|c|c}\toprule
    \textbf{Categories}  & Training &Testing&Recall(\%)&Precision(\%)\\
    \midrule
      Airplane: & 471 &  51 & 94.11 & 96\\ 
      Car:      & 257 &  31 & 93.33&96.55 \\
      Chair:    & 2508 & 281 &99.64 &98.93 \\
      Earphone: & 31  & 2&50 &25\\
      Guitar:   & 706 & 79&98.73 &98.73\\
      Lamp:     & 1086 & 115 &91.30&92.10 \\
      Pistol:   & 244 & 28 &100 &90.32\\
      Rocket:   & 51  & 6 & 100 & 66.66\\
      Skateboard:& 102 & 11&36.66&66.66\\
      Table:    & 1349 & 145 &96.55&97.90\\\bottomrule
      Total/ W. Average:    & 6806 & 749& 95.99&96.12\\\bottomrule
    \end{tabular}
    \label{table2}
    \end{subtable} 
     \begin{subtable}{\linewidth}
     
      \centering
    \caption{ Meta-SeL's confusion matrix result}
   
    \renewcommand{\arraystretch} {1}
    \hspace{-1cm}
    
    \begin{tabular}{ccccccccccc}\toprule
    \textbf{Categories}  & Airplane&Car& Chair& Earphone& Guitar& Lamp& Pistol&Rocket &Skateboard&Table\\
    \midrule
      Airplane: & 48 & 0 & 0 & 0 & 0 & 1 & 0 & 0&0 & 2 \\ 
      Car:      & 1&28 & 1&0 &1 &0 &0 &0 &0 &0  \\
      Chair:    &0 &0 &280 &0 &0 &0 &0 &0 &0 &1 \\
      Earphone:&0 &1 &0 &1 &0 &0 &0 &0 &0 &0 \\
      Guitar:   &0 &0 & 1& 0&78 &0 &0 &0 &0 &0 \\
      Lamp:     &6 &0 &0 &2 &0 &105 & 0& 0&2 &0 \\
      Pistol:   &0 &0 & 0& 0& 0& 0& 28&0 &0 &0 \\
      Rocket:  & 0& 0& 0& 1& 0&1 &0 &4 &0 &0 \\
      Skateboard:&0 &0 &0 &0 &0 &7 &0 &0 &4 &0 \\
      Table:    & 1& 0&1 &0 &0 &0 &3 &0 &0 &140 \\\bottomrule
    
    \end{tabular}
    \label{table3}
    \end{subtable}

\end{table*}

\section{Results and Discussion}
 
We evaluate Meta-SeL performance on a subset of the ShapeNet Core dataset. Proving that Meta-SeL yields promising results. Meta-SeL results are competitive with respect to other state-of-the-art methods.
We apply different filters to the dataset and elaborate on them. The experimental results show that Meta-SeL becomes resistant to normalization, random angle random rotations, translation, and added jittering noise.

\vspace{1pt}$\noindent\diamond$\textbf{Random shuffle-invariant, Sorted-invariant:}
We evaluate the performance of Meta-SeL on different orders of labels. We obtain the same result with different orders of labels including random shuffle, ascending sort, and descending sort. It proves that Meta-SeL is highly invariant to the order of input data. Additionally, it does not have bias \cite{shenavarmasouleh2019causes} towards classes with more instances over the ones with just a few instances.

  \vspace{1pt}$\noindent\diamond$\textbf{Normalization:} We normalize the base dataset into a unit sphere using the following steps \cite{qi2017pointnet}. First, we calculate the center by computing the mean of point clouds for each model. Then, we find the point which has the maximum distance from this center. Finally, we divide all points over this calculated maximum distance. By normalizing each model, the data is located between -1 and +1.

\vspace{1pt}$\noindent\diamond$\textbf{Rotation for 3D shapes:} 
We further evaluate Meta-SeL on a different angle with different rotations including the x-axis, y-axis, and z-axis. The result (accuracy$=88.38\%$) shows that we still have a better result than DGCNN. We obtain an accuracy of approximately five percent less than the base accuracy. It is noteworthy that we check which axis rotation has affected adversely to the accuracy. 

We test Meta-SeL with random angle rotation on x-axis. We randomly choose angles to rotate each model in train and test datasets. The result for \textit{N+RR(x-axis)} in Table \ref{table1} shows that the accuracy has decreased by approximately three percent in comparison with base accuracy. We further test Meta-SeL with random angle rotation on the y-axis. 
The result of this rotation states that the y-axis has the highest accuracy in comparison with other axis rotations. Additionally, we rotate the datasets using different angle rotations on the z-axis. The result shows that 3D models have the worst accuracy for z-axis rotation mode.



 \vspace{1pt}$\noindent\diamond$\textbf{Translation and rotation for 3D shapes:} 
 
 Further, we test Meta-SeL to check if it predicts as high as possible if translation happened to each 3D model or not.  We realize that a combination of normalization and translation has the highest accuracy. However, we test Meta-SeL on a combination of random angle random rotation and a translation. The result looks surprisingly expected due to the z-axis rotation and adding translation decreases its accuracy. 
  

  \vspace{1pt}$\noindent\diamond$\textbf{Jittering for 3D shapes:} 
  We add jittering to data to evaluate the performance of the Meta-SeL. Jittering is the process of adding random jittering noise to data to prevent overplotting in 3D models. Overplotting occurs when some 3D point clouds in the models overlap into a single point. Jittering can alleviate the problem and help Meta-SeL to generate an efficient projection learned model. Although jittering enables us to obtain a better result, adding translation due to jittering noise data decreases the accuracy. The final accuracy for this combination of jittering and translation is superior to the base result.
  

  \vspace{1pt}$\noindent\diamond$\textbf{Confusion Matrix:}
  We compute a confusion matrix as shown in table \ref{table3} for Meta-SeL evaluation on the Normalization dataset to tabulate how the work obtains competitive results. We show which category has the highest and lowest result. Table \ref{table3},\ref{tabel2} depicts that \textit{Chair} and \textit{Pistol} have the highest predicted statistic and recall. We notice that since \textit{Chair} and \textit{Table} might look similar to each other, one \textit{Chair's}  wrongly predicted sample goes to \textit{Table} and one of the wrongly predicted \textit{table} instances goes to \textit{Chair}. Furthermore, \textit{Airplane} wrongly labeled samples go to \textit{Table} and one of the wrongly predicted \textit{Table} instances goes to \textit{Airplane}. Additionally, two of the wrongly predicted \textit{Lamp} instances go to \textit{Skateboard} and all wrongly predicted labels for \textit{Skateboard} go to \textit{Lamp}. We observe that there is a trend among predicted labels. Whether Meta-SeL predicted correctly or wrongly, it predicts precisely.

\section{Conclusion}
 
In this work, we presented Meta-SeL for 3D point cloud classification on the ShapeNet core dataset. We explore each model to generate semantic space during training time. We calculated the projection function from vector space which is 3D point clouds models to the semantic space for both the training and testing process. Further, we employed a cosine similarity function to find the top match projection function among training and testing models. Then, we used an arg-max function to predict labels per each 3D model in the testing process. Additionally, we modified the input dataset with different orders, augmented data, and rotations to check the performance of Meta-SeL. The results indicate that Meta-SeL obtains competitive accuracy with an efficient time and cost.
 
\bibliographystyle{unsrt}  
\bibliography{references}
\end{document}